%% file: root.tex
\documentclass[letterpaper, 10 pt, conference]{ieeeconf}  % Comment this line out if you need a4paper

\IEEEoverridecommandlockouts                              % This command is only needed if 
\usepackage{etoolbox}
\makeatletter
\patchcmd{\@makecaption}
  {\scshape}
  {}
  {}
  {}
\makeatother
\usepackage{graphics} % for pdf, bitmapped graphics files
\usepackage{epsfig} % for postscript graphics files
\usepackage{amsmath} % assumes amsmath package installed
\usepackage{amssymb}  % assumes amsmath package installed
\usepackage{xcolor}
\usepackage{algorithm} %for algorithm box
\usepackage{algpseudocode}%for algorithm box
\usepackage{hyperref}

\title{\LARGE \bf
Preparation of Papers for IEEE Sponsored Conferences \& Symposia*
}

\title{\LARGE \bf
Learning Switching Linear Dynamical System Parameters for Differentiable Optimal Control}

\title{\LARGE \bf
Learning Contact Dynamics and Cost Functions for Multi-Modal Differentiable Optimal Control}

\title{\LARGE \bf
Learning Switching Linear Dynamical System Parameters for Differentiable Reactive and Predictive Control}

\title{\LARGE \bf
Learning Differentiable Reactive and Predictive Controllers for Switching Linear Dynamical Systems}

\title{\LARGE \bf
Learning Reactive and Predictive Differentiable Controllers for Switching Linear Dynamical Models}

\author{Saumya Saxena$^{1}$, Alex LaGrassa$^{1}$ and Oliver Kroemer$^{1}$% <-this % stops a space
\thanks{*This work was in part supported by the National Science Foundation under Grant No. CMMI-1925130 and IIS-1956163. Any opinions, findings, and conclusions or recommendations expressed in this material are those of the author(s) and do not necessarily reflect the views of the NSF.}% <-this % stops a space
\thanks{$^{1}$Robotics Institute, School of Computer Science, Carnegie Mellon University, Pittsburgh PA 15123, USA {\tt\small \{saumyas, alagrass, okroemer\}@andrew.cmu.edu}}%
}

\begin{document}

\maketitle
\thispagestyle{empty}
\pagestyle{empty}
%%%%%%%%%%%%%%%%%%%%%%%%%%%
\newcommand{\Crhorho}{\mathbf{C}_{\boldsymbol{\rho}_t, \boldsymbol{\rho}_{t-1}}}

%%%%%%%%%%%%%%%%%%%%%%%%%%%%%%%%%%%%%%%%%%%%%%%%%%%%%%%%%%%%%%%%%%%%%%%%%%%%%%%%
\input{abstract}
\input{introduction}
\input{relatedwork}
\input{method}

\input{experiments}

\section{CONCLUSIONS}
% \vspace{-0.3cm}
\input{conclusion}

\section*{ACKNOWLEDGMENT}
\footnotesize The authors would like to thank Kevin Zhang, Blake Buchanan and Jacky Liang for their support during this work. The authors also would like to thank Prof. Zico Kolter for engaging technical discussions on differentiable control algorithms that form the basis of this work.

\bibliographystyle{IEEEtran}
\bibliography{references}
% \section*{}
\input{appendix}
\end{document}

%% file: abstract.tex
\begin{abstract}

Humans leverage the dynamics of the environment and their own bodies to accomplish challenging tasks such as grasping an object while walking past it or pushing off a wall to turn a corner. Such tasks often involve switching dynamics as the robot makes and breaks contact. Learning these dynamics is a challenging problem and prone to model inaccuracies, especially near contact regions. In this work, we present a framework for learning composite dynamical behaviors from expert demonstrations. We learn a switching linear dynamical model with contacts encoded in switching conditions as a close approximation of our system dynamics. We then use discrete-time LQR as the differentiable policy class for data-efficient learning of control to develop a control strategy that operates over multiple dynamical modes and takes into account discontinuities due to contact. In addition to \textit{predicting} interactions with the environment, our policy effectively \textit{reacts} to inaccurate predictions such as unanticipated contacts.
Through simulation and real world experiments, we demonstrate generalization of learned behaviors to different scenarios and robustness to model inaccuracies during execution.

\end{abstract}

%% file: introduction.tex
\section{INTRODUCTION}
Everyday human tasks involve intricate interactions with the environment, manifesting in changing dynamics and impacts. For example, to pick up a bottle in one smooth motion while walking past it, a human first slows down, comes in contact, and then applies an appropriate force to pick it up without toppling it over. Modeling the dynamics for such tasks is a challenging problem. Even small inaccuracies in the learned model, especially near the contact regions, can cause catastrophic failures and unstable behaviour. One approach to prevent such problems is to not rely on the learned models entirely \cite{nagabandi2018neural, janner2019trust}, rather use them as close approximations of the system’s behaviour, and at the same time design/learn a controller that is both robust to the model inaccuracies and stable under execution noise. 

In this work, we develop a control strategy that operates over multiple dynamical modes, takes into account discontinuities due to contact, and is robust to inaccuracies in the learned dynamics model. Our approach focuses on learning the dynamic behavior of an expert and then generalizing that behavior to unseen scenarios. 

Systems for which the dynamics change discontinuously, usually after contact is made or broken, can be modeled as \textit{switching linear dynamical systems} (SLDS) \cite{kroemer2015towards, becker2019switching,linderman2016recurrent, dabrowski2018naive,toussaint2005learning, lee2017unsupervised}. The set of discrete dynamics in a SLDS are governed by hidden states called \textit{modes}. We model impacts as conditions governing the switch between these modes, which incorporates discontinuities into our dynamics model. We use system identification to learn the parameters of the SLDS including the impact dynamics and mode prediction function.

Taking inspiration from recent approaches that add prior structure to policy architecture \cite{amos2018differentiable, karkus2019differentiable, toussaint2018differentiable} for data-efficient learning of control, we use discrete-time LQR as our differentiable policy class. In addition to being sample efficient, this policy class provides a feedback control scheme that captures expert behavior in the learned cost/value functions which can be used to generalize to other scenarios. For our complex task modeled as a SLDS, we use a single LQR policy and parameterize it using a different cost function and goal condition for each mode. Using cost matrices learned from a single scenario, we can generalize to different goal conditions while achieving the same overall behavior. 

Forward propagating the control using the learned model allows us to predict mode switches and contacts by extension. To make the controller robust to inaccurate mode predictions, we combine our method with a reactive control scheme \cite{sugimoto2012emosaic}, to switch modes upon \textit{observing} inaccurate predictions of state transitions. We use model predictive control during execution to replan using the observed modes.

The contributions of this paper are twofold: 1) a method for modeling and learning switching contact-based linear dynamics and using it for efficiently learning a differentiable feedback controller that operates over multiple dynamical modes, takes into account discontinuities due to contact, and generalizes it to unseen scenarios, 2) a control scheme that is robust to inaccuracies in the learned dynamics model and during execution can predict, as well as react to, unanticipated dynamic contact events. An overview of the method and results can be found in the supplementary video \href{https://drive.google.com/file/d/1_wCi3WS61ybEyUHivgc2DINPatjorR_o/view?usp=sharing}{(link)}.

%% file: relatedwork.tex
\begin{figure*}[t]
    \begin{center}
    \includegraphics[width=\textwidth]{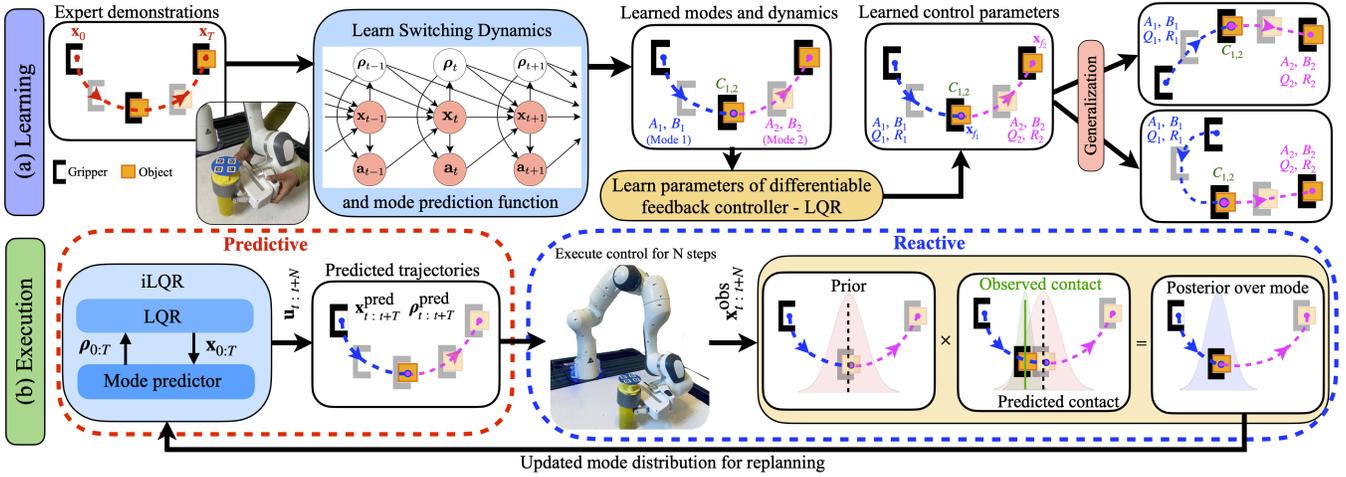}
    \vspace{-0.7cm}
    \caption{Training and execution pipeline (a) Given a set of expert demonstrations, learn the parameters of a switching linear dynamical system (transition dynamics, contact dynamics and mode prediction function) using system identification. Using the learned dynamics parameters and the expert demonstrations, learn a differentiable feedback controller (LQR), parameterized using a different cost function and goal condition for each mode, using imitation learning. Learning the cost matrices allows us to capture the behavior of the expert in each mode, thus enabling the learned controller to generalize to unseen scenarios. (b) During execution, we use iLQR to iteratively evaluate the modes along a trajectory, and use these modes and hence the corresponding dynamics, impact and cost parameters to solve the LQR problem. To make the controller robust to inaccuracies in the learned  model, we use MPC to combine our predictive controller with a reactive control scheme. The predictive part adapts the control in anticipation of mode switches, and the reactive part observes unanticipated mode transitions, updates the current mode and replans.}
    \label{fig:flowchart}
    \end{center}
    \vspace{-0.8cm}
\end{figure*}
\section{RELATED WORK}
Contacts and changing dynamics are ubiquitous in manipulation \cite{kroemer2019review} and collaborative tasks. Model-based control approaches for learning contact-rich manipulation have been proposed \cite{gu2016continuous}, \cite{mishra2017prediction}, but their asymptotic performance suffers due to model-bias. To alleviate this problem, other approaches combine model-based and model-free methods  \cite{nagabandi2018neural, sergey2015learning}. Our approach relies on using a robust control scheme \cite{dullerud2013course} to deal with inaccuracies in the learned dynamics model.\\
\indent Modeling tasks with non-stationary dynamics as switching linear dynamical systems (SLDS) is a well known problem. There are approaches that do \cite{toussaint2005learning, lee2017unsupervised} and do not \cite{kroemer2015towards, becker2019switching,linderman2016recurrent, dabrowski2018naive} model discontinuous mode transitions. Some other approaches for modelling contacts and learning contact based dynamics include \cite{pfrommer2020contactnets, fazeli2017parameter, calandra2015learning, khader2020data}. Our approach for modeling contact as mode switches compares best with \cite{fazeli2017learning}, where the mapping from pre-impact velocity to post-impact velocity is learned. %These methods effectively model contacts and dynamic mode switches, but do not discuss how these models can be used for control.%
Methods such as \cite{kroemer2015towards, manschitz2015learning} learn controllers for each mode separately, and switch between controllers using the learned predictive guard conditions. Such methods are susceptible to failures due to inaccuracies in these learned conditions. Methods such as eMOSAIC \cite{sugimoto2012emosaic}, on the other hand, do not learn a predictive model for mode switching, but switch modes and control reactively upon observed changes in the effects of actions.  In our method, we combine the advantages of both the predictive and reactive control schemes to develop a robust controller that acts over multiple discrete modes and contact.

Model-free approaches \cite{andrychowicz2020learning, mnih2015human, gu2016q} obviate the need for learning the dynamics, but face challenges in terms of sample complexity. Recent approaches to alleviate these challenges propose adding prior structure to the policy architecture \cite{amos2018differentiable, karkus2019differentiable, toussaint2018differentiable}. We take inspiration from this work and use discrete-time LQR as our differentiable policy class.  

%% file: method.tex
\vspace{-0.05cm}
\section{METHOD}
% \vspace{-0.5cm}
In this section we first elaborate on our method for modeling and learning parameters of the switching dynamical system using system identification. We then model the controller and learn the control parameters using imitation learning. Finally we introduce our robust control scheme for executing the learned policy.
\subsection{Modeling the Switching Linear Dynamical System}
Graphical model for our switching dynamical system is shown in the second panel in Fig. \ref{fig:flowchart}(a). State of the system at time \textit{t} is given by $\mathbf{x}_t = [\mathbf{q}_t, \mathbf{\dot{q}}_t]$ where $\mathbf{x}_t \in \mathbb{R}^n$, $\mathbf{q}_t$, $\mathbf{\dot{q}}_t$ are configuration and velocities respectively. State is composed of state of the robot and states of objects of interest in the environment. At every state $\mathbf{x}_t$ the robot takes an action $\mathbf{u}_t \in \mathbb{R}^m$ and transitions to state $\mathbf{x}_{t+1}$. The transition dynamics and the control parameters at every time step are dependent on the discrete \textit{mode} $\boldsymbol{\rho}_t$, which is a hidden variable. 

\subsubsection{Mode Prediction}
\label{sec:mode_classification}
We assume that the current mode $\boldsymbol{\rho}_t$ is only dependent upon the current state $\mathbf{x}_t$. We call the function for mode prediction a \textit{classifier} $\boldsymbol{\rho}_t = M(\mathbf{x}_t).$ The classifier is modeled as a fully-connected neural network that takes as input current state $\mathbf{x}_t$ and outputs a one-hot vector over modes. For a picking task the state $\mathbf{x}_t$ will include the position and orientation of the object we want to pick up. The classifier can learn to predict modes that depend upon more general conditions such as relative position between the gripper and the object, as used in our model. We always set the number of modes in the model to be greater or equal to the actual number of modes required by the task. The number of learned modes can be less than or equal to the maximum value specified.

\subsubsection{Transition and Interaction dynamics}
The transition dynamics are modeled as follows: 
    % \vspace{-0.2cm}
\begin{align}
    \mathbf{x}_{t+1} &= \mathbf{A}_{\boldsymbol{\rho}_t} \mathbf{C}_{\boldsymbol{\rho}_t, \boldsymbol{\rho}_{t-1}} \mathbf{x}_t + \mathbf{B}_{\boldsymbol{\rho}_t} \mathbf{u}_t
    \label{eq:dyn}
\end{align}
where $\mathbf{A}_{\boldsymbol{\rho}_t} \in \mathbb{R}^{n \times n}$ and $\mathbf{B}_{\boldsymbol{\rho}_t} \in \mathbb{R}^{n \times m}$ correspond to the discrete mode $\boldsymbol{\rho}_t$ at time t. Thus, for a set of $N$ modes, there will be $N$ such transition matrices that we need to learn. The matrix $\mathbf{C}_{\boldsymbol{\rho}_t, \boldsymbol{\rho}_{t-1}} \in \mathbb{R}^{n \times n} $ represents the impact or interaction dynamics which depends upon the mode switch between the current and previous time step. For example, in case of a pushing task with two modes, the first mode is when the robot is in free motion and the second mode is when it is in contact with the light object and is pushing it. An impact occurs between the robot and the object at the instant when contact is made, that is, when the mode \textit{switches} from mode 1 to 2. Upon impact, the velocity of a system changes instantaneously while the position remains the same, which can be modeled as momentum transfer between the robot and the object. This is the source of discontinuity in the dynamics which we model here explicitly.

We take inspiration from approaches that model contact as a mapping from pre-impact to post-impact velocity \cite{fazeli2017learning}. Consider a simple system where the gripper and object states are given as $\mathbf{x}^{g}_t = [\mathbf{q}^{g}_t, \mathbf{\dot{q}}^{g}_t]$ and $\mathbf{x}^{o}_t = [\mathbf{q}^{o}_t, \mathbf{\dot{q}}^{o}_t]$, respectively. The impact dynamics for this system can be written as
    % \vspace{-0.2cm}
\begin{align*}
    \begin{bmatrix}
    \mathbf{q}^{g}_{t^+}\\
    \dot{\mathbf{q}}^{g}_{t^+}\\
    \mathbf{q}^{o}_{t^+}\\
    \dot{\mathbf{q}}^{o}_{t^+}
    \end{bmatrix} 
    = 
    \underbrace{ \begin{bmatrix}
    1 & 0 & 0 & 0\\
    0 &  \frac{m_g-em_o}{m_g+m_o} & 0 &  \frac{(1+e) m_o}{m_g+m_o}\\
    0 & 0 & 1 & 0\\
    0 &  \frac{(1+e) m_g}{m_g+m_o} & 0 &  \frac{m_o -e m_g}{m_g+m_o}
    \end{bmatrix} }_{\mathbf{C}}
    \begin{bmatrix}
    \mathbf{q}^{g}_{t^-}\\
    \dot{\mathbf{q}}^{g}_{t^-}\\
    \mathbf{q}^{o}_{t^-}\\
    \dot{\mathbf{q}}^{o}_{t^-}
    \end{bmatrix},
\end{align*}
    % \vspace{-0.3cm}
where $e$ is the coefficient of restitution between the gripper and the object, $m_g$ and $m_o$ are the mass matrices for the gripper and the object respectively, and $\mathbf{q}_{t^-}, \dot{\mathbf{q}}_{t^-}$ and $\mathbf{q}_{t^+}, \dot{\mathbf{q}}_{t^+}$ are the position and velocity before and after impact respectively.
We observe that some properties of this  impact/interaction matrix $\mathbf{C}$ make it easier to learn from demonstration and apply it in a wide range of applicable scenarios. First,  $\mathbf{C}$ is only dependent on the system properties and not on the velocity or position. Thus, we can parameterize and learn it using expert demonstrations and then use it in different scenarios. Second, we need only learn the parameters corresponding to changes in velocity because the position remains the same during impact. Finally, even though we model a different impact matrix for each possible mode switch, we end up learning impact matrices only for the mode switches that are observed during demonstrations, pruning the number of impact matrices that need to be learned. To ensure that the impact only occurs upon a mode \textit{switch}, we write the impact dynamics as
$\mathbf{C}_{\boldsymbol{\rho}_{t-1}, \boldsymbol{\rho}_t} = \boldsymbol{\rho}_{t-1}^{\top} \boldsymbol{\rho}_t \mathbf{I} + (1-\boldsymbol{\rho}_{t-1}^{\top} \boldsymbol{\rho}_t ) \mathbf{C}$
because when $\boldsymbol{\rho}_{t-1}= \boldsymbol{\rho}_t$, $\mathbf{C}_{\boldsymbol{\rho}_{t-1}, \boldsymbol{\rho}_t} = \mathbf{I}$ and when $\boldsymbol{\rho}_{t-1} \neq \boldsymbol{\rho}_t$, $\mathbf{C}_{\boldsymbol{\rho}_{t-1}, \boldsymbol{\rho}_t} = \mathbf{C}$ where $\mathbf{C}$ is learned from demonstrations.

\subsection{Learning the Switching Dynamics Model}
\label{sec:sysID_learning}
Having defined the structure of the dynamical system, we move on to learning the model parameters from expert demonstrations using system identification. The expert demonstrations are given as a set of trajectories $ \{\mathbf{x}^*_t, \mathbf{u}^*_t, \mathbf{x}^*_{t+1}\}_{t = \{0, \dots, T-1\}}$ where T is the total length of the trajectory. The system identification cost is written as:\begin{gather*}
    \boldsymbol{\rho_t} = M(\mathbf{x}_t^*), \;\;
    \mathbf{x}_{t+1}^\text{pred} = \mathbf{A}_{\boldsymbol{\rho}_t} \mathbf{C}_{\boldsymbol{\rho}_t, \boldsymbol{\rho}_{t-1}} \mathbf{x}_t^* + \mathbf{B}_{\boldsymbol{\rho}_t} \mathbf{u}_t^*\\
    \mathcal{L}_\text{SysID} = ||\mathbf{x}_{1:T}^* - \mathbf{x}_{1:T}^\text{pred}||_2.
\end{gather*}
Taking gradient with respect to the above cost we together learn $\mathbf{A}_{\boldsymbol{\rho}_t}, \mathbf{B}_{\boldsymbol{\rho}_t}, \mathbf{C}_{\boldsymbol{\rho}_t, \boldsymbol{\rho}_{t-1}}$, and the classifier $M(\cdot)$.

\subsection{Modeling the Differentiable Policy}
Having learned the system dynamic parameters, we shift our focus to modeling a controller that can perform the task demonstrated by the expert. More importantly, the controller should
1) robustly perform the task, 2) match the behavior of the expert through the learned cost function, such as slowing down before grasping, 3) generalize to other task instances that use the same objects. We design our controller as a discrete-time linear quadratic regulator (LQR), a differentiable feedback controller that can be represented as the solution to the following optimization problem:
\begin{equation}
\label{eq:LQR}
\begin{aligned}
& \underset{\mathbf{u}_t}{\text{min}}
& & \!\!\!\!\sum_{t=0}^T \big( (\mathbf{x}_t-\mathbf{x}_{f_{\boldsymbol{\rho}_t}})^{\top}\mathbf{Q}_{\boldsymbol{\rho}_t} (\mathbf{x}_t-\mathbf{x}_{f_{{\boldsymbol{\rho}_t}}})+ \mathbf{u}_t^{\top}\mathbf{R}_{\boldsymbol{\rho}_t} \mathbf{u}_t \big)\\
& \text{s.t.}
& & \!\!\!\!\mathbf{x}_{t+1} = \mathbf{A}_{\boldsymbol{\rho}_t} \mathbf{C}_{\boldsymbol{\rho}_t, \boldsymbol{\rho}_{t-1}} \mathbf{x}_t + \mathbf{B}_{\boldsymbol{\rho}_t} \mathbf{u}_t, \; i = 1, \ldots, T\\
&&& \!\!\!\!\mathbf{x}_0 = \mathbf{x}_\text{init}
\end{aligned}
\end{equation}
where $\mathbf{x}_{f_{\boldsymbol{\rho}_t}}, \mathbf{Q}_{\boldsymbol{\rho}_t}, \mathbf{R}_{\boldsymbol{\rho}_t}$ correspond to the goal configuration and cost matrices for mode $\boldsymbol{\rho}_t$.  This formulation allows us to capture the behavior of the system in each mode using the cost matrices $\mathbf{Q}$ and $\mathbf{R}$ for a corresponding goal location $\mathbf{x}_f$. It is important to note that the behavior over the entire task is governed by the sequence of behaviors of the learned modes. Thus, once learned from expert demonstrations, as illustrated in Section \ref{sec:sysID_learning}, the \textit{sequence} of modes should be kept fixed to successfully complete the task. 

Using LQR feedback controller enables optimal responses to anticipated changes in dynamics and impact conditions in the trajectory. By learning cost matrices, the agent is essentially learning a value function or reward structure around each goal configuration. Using learned cost matrices allows the controller to generalize to different sub-goal configurations along the trajectory. Keeping the sequence of modes, and thus the \textit{behavior} along the trajectory, fixed, we can shift around the learned sub-goal locations $(\mathbf{x}_f)$ to new desired locations and still attain the same overall behavior.

\subsection{Learning the Control Parameters}
\label{sec:control_param_learning}
As illustrated in Section \ref{sec:sysID_learning}, for each expert demonstration, we have already learned the mode prediction $\boldsymbol{\rho_t} = M(\mathbf{x}_t^*)$ and the corresponding dynamic parameters $\mathbf{A}_{\boldsymbol{\rho}_t}, \mathbf{B}_{\boldsymbol{\rho}_t}, \mathbf{C}_{\boldsymbol{\rho}_t, \boldsymbol{\rho}_{t-1}}$ for each time step. Now, we use imitation learning to learn the control parameters $\mathbf{x}_{f_{\boldsymbol{\rho}}}, \mathbf{Q}_{\boldsymbol{\rho}}$ and $\mathbf{R}_{\boldsymbol{\rho}}$. For each expert initial condition $ \{\mathbf{x}^*_0\}_{t = \{0, \dots, T\}}$, using the corresponding learned dynamic parameters and modes, we solve the optimization problem in \eqref{eq:LQR} using discrete-time Riccati equations \cite{bemporad2002explicit}. Let the solution to the LQR problem for each initial state $\mathbf{x}^*_0$ be given as
\begin{align*}
    \mathbf{x}^\text{LQR}_{0:T},\! \mathbf{u}^\text{LQR}_{0:T-1} \!=\! LQR(\mathbf{x}^*_0,\! \mathbf{A}_{\boldsymbol{\rho}_t},\! \mathbf{B}_{\boldsymbol{\rho}_t},\! \mathbf{C}_{\boldsymbol{\rho}_t, \boldsymbol{\rho}_{t-1}},\! \mathbf{Q}_{\boldsymbol{\rho}_t},\! \mathbf{R}_{\boldsymbol{\rho}_t},\! \mathbf{x}_{f_{\boldsymbol{\rho}_t}}).
\end{align*}
The imitation cost can then be written as $$\mathcal{L}_\text{Imitation} = ||\mathbf{x}_{0:T}^* - \mathbf{x}_{0:T}^\text{LQR}||_2 + ||\mathbf{u}_{0:T-1}^* - \mathbf{u}_{0:T-1}^\text{LQR}||_2.$$
The differentiabilty of the LQR controller allows us to learn the goal configurations $\mathbf{x}_{f_{\boldsymbol{\rho}}}$ and cost functions $ \mathbf{Q}_{\boldsymbol{\rho}}, \mathbf{R}_{\boldsymbol{\rho}}$ corresponding to each mode by minimizing the above cost.

\subsection{Execution}

We have shown how we learn the dynamics model and controller from expert demonstrations. Now we move on to how we can solve similar tasks with new initial configurations of the robot and objects in the environment using these learned parameters. We also show how the controller accounts for inaccuracies in the learned dynamics model. 

\subsubsection{Iterative Linear Quadratic Regulator}
% \vspace{
\begin{algorithm}
\caption{Iterative LQR}
\begin{algorithmic}
\State \scalebox{1.0}{$(i)$ represents the iteration number}
\State \scalebox{1.0}{T is the length of the trajectory}
\State \scalebox{1.0}{$j,k \in \{ 1, \dots, N\}$ where $N$ is the number of modes}
\State \textbf{Given}: Initial state $\mathbf{x}_{init}$. Initialize modes $\boldsymbol{\rho}_{0:T}$
\State \textbf{Parameters}: \scalebox{0.9}{$M(.),\mathbf{A}_{\boldsymbol{\rho}_j}, \mathbf{B}_{\boldsymbol{\rho}_j}, \mathbf{C}_{\boldsymbol{\rho}_j, \boldsymbol{\rho}_{k}}, \mathbf{x}_{f_{\boldsymbol{\rho}_j}}, \mathbf{Q}_{\boldsymbol{\rho}_j}, \mathbf{R}_{\boldsymbol{\rho}_j} $}
% \begin{algorithmic}
    % \Procedure{Train}{$x_0, A, B$}
    %   \scriptsize
\Procedure{iLQR}{Parameters}
    \While{not converged}
        \State \!\!\!\!\scalebox{0.83}{$\mathbf{x}^{(i)}_{0:T},\mathbf{K}_{0:T-1}$ $\gets$ LQR($\mathbf{A}_{\boldsymbol{\rho}_t^{(i)}}, \!\mathbf{B}_{\boldsymbol{\rho}_t^{(i)}}, \!\mathbf{C}_{\boldsymbol{\rho}_t^{(i)}, \boldsymbol{\rho}_{t-1}^{(i)}}\!, \!\mathbf{x}_{f_{\boldsymbol{\rho}_t^{(i)}}}, \!\mathbf{Q}_{\boldsymbol{\rho}_t^{(i)}}, \!\mathbf{R}_{\boldsymbol{\rho}_t^{(i)}}$)}
        \State \!\!\!\!\scalebox{0.83}{$\boldsymbol{\rho}_t^{(i+1)} = M(\mathbf{x}_t^{(i)})$} for all $t$
        \State \!\!\!\!\scalebox{0.83}{converged $\gets$ $||\mathbf{x}_{0:T}^{(i+1)} - \mathbf{x}_{0:T}^{(i)}||_2 + ||\mathbf{u}_{0:T-1}^{(i+1)} - \mathbf{u}_{0:T-1}^{(i)}||_2$ is small}
    \EndWhile
\EndProcedure
\end{algorithmic}
\label{alg:ilqr}
\end{algorithm}

We formulate the problem of evaluating the control sequence for tasks involving multiple discrete dynamical modes as an iterative LQR problem. The approach is very similar to a standard iLQR problem where, starting with an initial trajectory, we linearize the dynamics around the trajectory and use these dynamics to solve the optimal LQR problem iteratively. In our approach, as illustrated in Algorithm \ref{alg:ilqr}, at every iteration, we evaluate the mode for each step of the trajectory and populate the dynamics, impact and cost matrices accordingly to iteratively solve the LQR problem. It is important to note here that, from the expert demonstrations, we learned the cost matrices that capture the behavior of the system while achieving goal states $\mathbf{x}_f{_{\boldsymbol{\rho}_j}}$. By substituting  $\mathbf{x}_f{_{\boldsymbol{\rho}_j}}$ with new desired goal states, we can generalize to different initial and goal configurations for a task with the same sequence of modes. A high-level planner can be used to rearrange the modes to generalize to similar tasks with different mode sequences, but this has been left for future work.

\subsubsection{Model Predictive Control}
\begin{algorithm}
\caption{Model Predictive Control}
\begin{algorithmic}
\State $T_{LQR}$ is time horizon for execution before replanning
\State $N_{MPC} = T/T_{LQR}$
\State $T$ is the length of the trajectory
\State $t$ is the current time step

\State $t \gets 0$
\For{n in $N_{mpc}$}
    \State \!\!\!\!\scalebox{1.0}{$\mathbf{K}_{t:T}$ $\gets$ iLQR($\mathbf{x}_t,\mathbf{A}_{\boldsymbol{\rho}}, \!\mathbf{B}_{\boldsymbol{\rho}}, \!\mathbf{C}_{\boldsymbol{\rho}, \boldsymbol{\rho}}\!,
    \! \mathbf{R}_{\boldsymbol{\rho}},
    \!\mathbf{Q}_{\boldsymbol{\rho}},
    \!\mathbf{x}_{f_{\boldsymbol{\rho}},T\!-\!t}$)}
    \For{i in $T_{LQR}$}
        \State $\boldsymbol{\rho}_t^{post} \gets \text{Posterior}(\mathbf{x}_{t},\mathbf{x}_{t-1},\mathbf{a}_{t-1})$ \Comment{\eqref{eq:posterior}}
        \State $\mathbf{a}_t \gets \mathbf{K}_t(\mathbf{x}_t - \mathbf{x}_{f_{\boldsymbol\rho_t}}^{post})$
        \State $\mathbf{x}_{t+1} \gets \text{step}(\mathbf{a}_t)$
        \State $t \gets t_{t+1}$
    \EndFor
\EndFor
\end{algorithmic}
\label{alg:mpc}
\end{algorithm}
% \vspace{-2cm}
Since LQR is a feedback controller, it is robust to small disturbances and noise that are encountered during execution. However, our controller should be robust to inaccuracies in our learned model, especially the learned mode predictor $M(\mathbf{x}_t)$. Consider a scenario where for a grasping task, the classifier learns that contact occurs when the measured relative position between the gripper and the object is less than some threshold value. During execution, contact could potentially occur a little before, or a little after, this threshold is crossed. If the threshold is a little too large and the next mode requires the robot to move away from the object location, the robot might start moving away before actually grasping the object. To make the system robust to such scenarios, we use model predictive control (MPC), Algorithm \ref{alg:mpc}, to \textit{replan} using observations from the environment. We evaluate a  posterior over the mode estimate using the current observation as 
\begin{align}
\label{eq:posterior}
p(\boldsymbol{\rho}_t|\mathbf{x}_{t},\! \mathbf{u}_{t-1}, \!\mathbf{x}_{t-1}) \!=\! z * p(\boldsymbol{\rho}_t| \mathbf{x}_{t}) p(\mathbf{x}_{t}| \mathbf{x}_{t-1},\! \mathbf{u}_{t-1},\! \boldsymbol{\rho}_t)
\end{align}
where $p(\mathbf{x}_{t}| \mathbf{x}_{t-1}, \mathbf{u}_{t-1}, \boldsymbol{\rho}_t)$ gives the probability of seeing the observed transition given a certain mode $\boldsymbol{\rho}_t$, $p(\boldsymbol{\rho}_t| \mathbf{x}_{t})$ is the predicted mode from the learned classifier, and $z$ is the normalization constant.
This approach to evaluating the mode using observed state transitions is similar to the eMOSAIC approach~\cite{sugimoto2012emosaic}. The above posterior can also be calculated using measured contacts if the robot is equipped with force/contact sensing. Using this posterior estimate of the mode, we replan as given in Algorithm \ref{alg:mpc}. Adding this framework to our approach, we enable our controller to be predictive of mode transitions (using the classifier) as well as reactive (using measurements to evaluate posterior) to unanticipated mode changes and impacts conditions making the controller robust to learned model inaccuracies.

%% file: experiments.tex
\vspace{0.4cm}
\begin{figure}[t]
    \centering
    \includegraphics[scale=0.15, trim={0 0.cm 0 0cm}, clip]{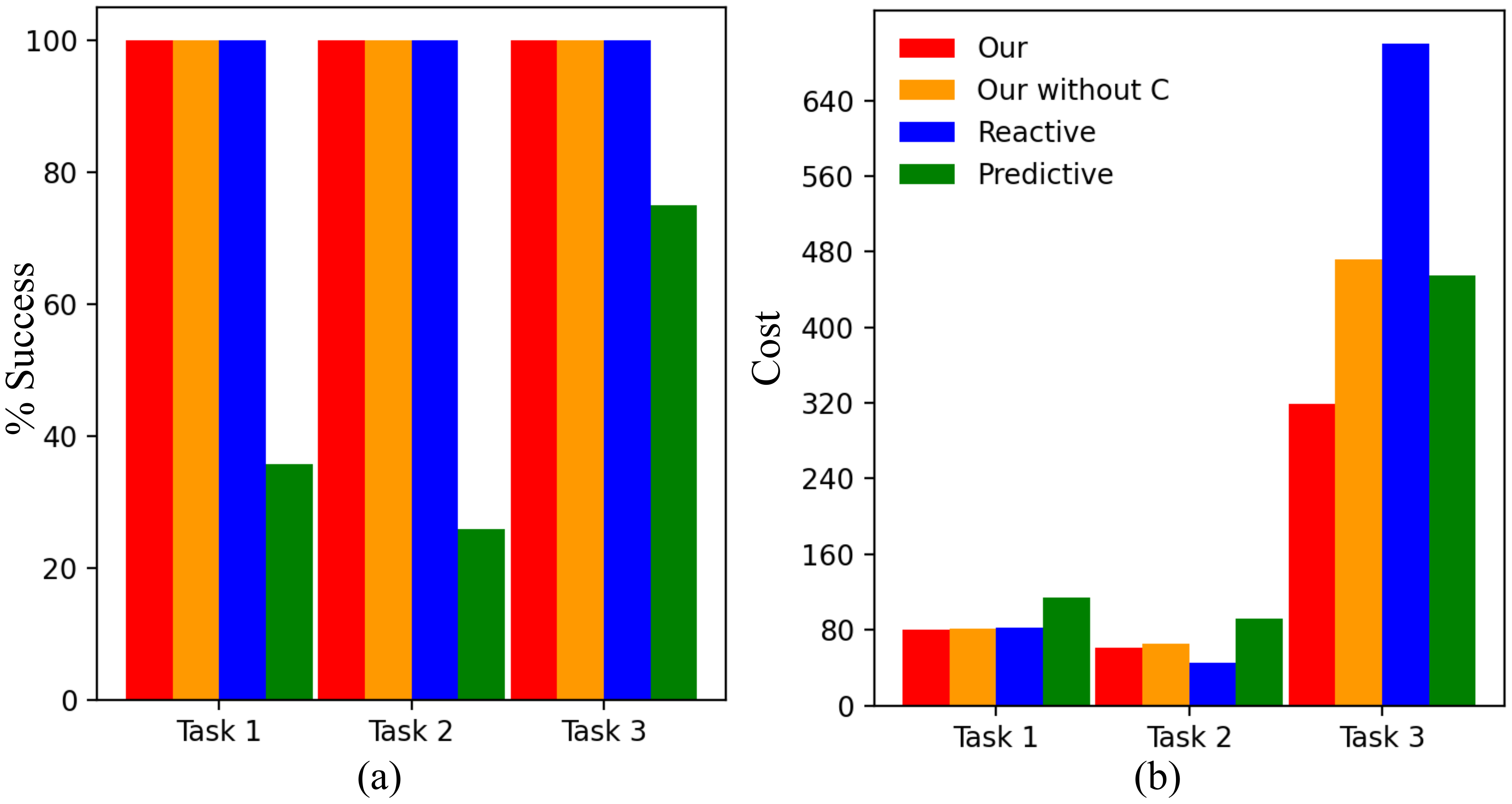}
    \vspace{-0.4cm}
    \caption{(a) Success rate (b) Execution cost for the 3 Box2D tasks}
    \label{fig:success_rate}
    \vspace{-0.4cm}
\end{figure}
\vspace{-0.3cm}
\section{EXPERIMENTS}
\vspace{0.3cm}
\begin{figure*}
    \centering
    \includegraphics[width=0.95\textwidth,trim={0cm 0cm 0cm 0cm},clip]{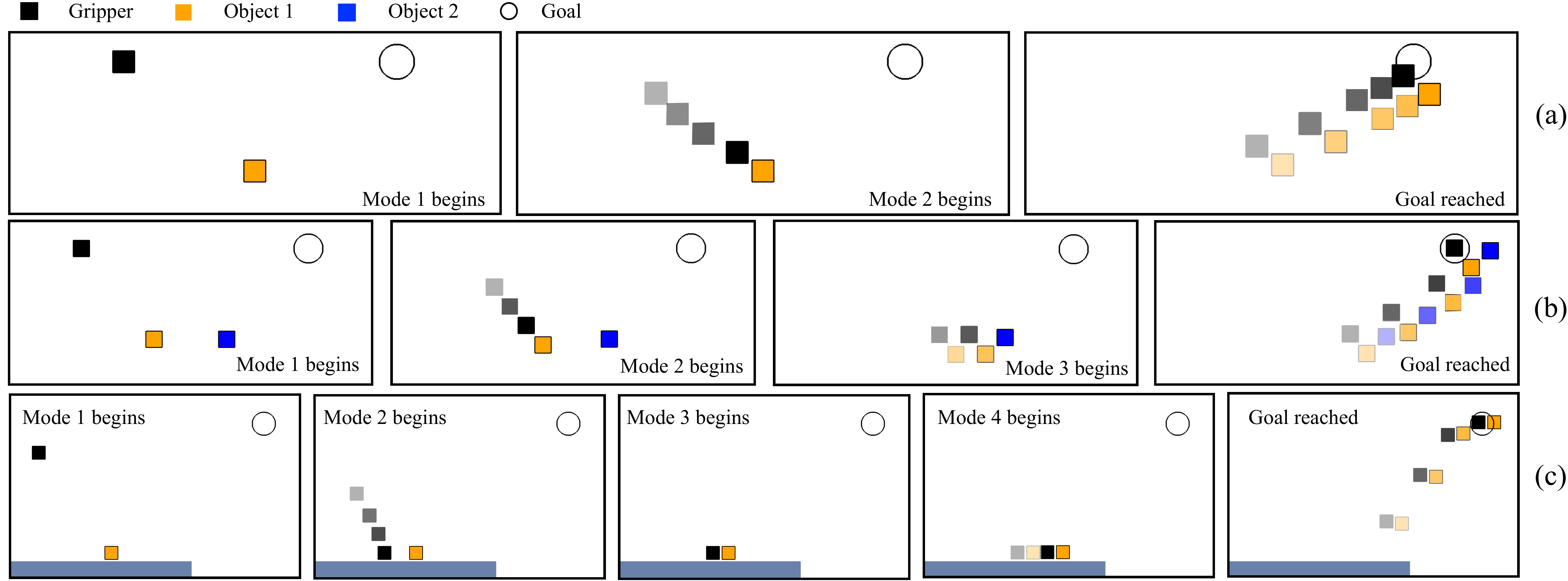}
    \vspace{-0.3cm}
    \caption{Learned behaviors for (a) 1-object pickup task, (b) 2-object pickup task, and (c) pickup task requiring sliding along the table to align with the object before picking it up, A 2D gripper (black block) picks up objects (orange and blue blocks) and takes it to goal region (black circle). Motion of gripper and objects is depicted by overlayed transparent frames.}
    \label{fig:sampleTrajs}
    \vspace{-0.5cm}
\end{figure*}

The proposed method was evaluated in simulation using the Box2D environment and in the real-world on the 7DOF Franka-Emika Panda robot arm.

% \label{table:learned_params}
% \vspace{-0.3cm}
\begin{table}[t]
\caption{RMSE of learned masses and learned goal configurations.}
\vspace{-0.6cm}
\begin{center}
\resizebox{0.4\textwidth}{!}{%
\begin{tabular}{ | c | c | c | } 
\hline
 & Masses (kg) & Goal configurations (m)  \\
\hline
Task 1 & 0.029 & 0.095 \\
\hline
Task 2 & 0.034 & 0.088 \\ 
\hline
Task 3 & 0.009 & 0.026 \\
\hline
\end{tabular}}
\end{center}
\label{tab:learned_params}
\vspace{-1.0cm}
\end{table}

\subsection{Simulation Experiments}
To evaluate our controller, which is both reactive and predictive and takes into account discontinuities due to contact, we compare our results to 1) a purely reactive eMOSAIC controller \cite{sugimoto2012emosaic}, 2) a purely predictive version of our controller, i.e. without using the posterior over the mode to replan using MPC, and 3) a predictive and reactive controller that doesn't takes into account discontinuities due to contact. Experiments were designed to test the robustness of these controllers to process noise, model inaccuracies and variations in shape geometry and initial configurations not seen during training. We use two metrics to compare results: 1) success rate - defined by whether the robot is able to pick up the object(s) of interest and take them to a goal region in a specified time 2) cost to perform the task, defined as the LQR cost:
$ \sum_{t=0}^T \big( (\mathbf{x}_t-\mathbf{x}_{f_{\boldsymbol{\rho}_t}})^{\top}\mathbf{Q}_{\boldsymbol{\rho}_t} (\mathbf{x}_t-\mathbf{x}_{f_{\boldsymbol{\rho}_t}})+ \mathbf{u}_t^{\top}\mathbf{R}_{\boldsymbol{\rho}_t} \mathbf{u}_t \big)$. We perform three tasks in simulation: 1) pick up one object and deliver to goal location 2) pick up two objects and deliver to goal location 3) pick up task while interacting with the ground to align with the object.\\
\indent Expert demonstrations are collected by sampling the initial configuration for the gripper and fixed initial configurations for the objects. We use ground truth dynamic properties of the objects and manually defined cost functions to collect expert demonstrations for each task in simulation. By hand-tuning the cost matrices, we can control how much the gripper slowed down before picking up each object. For each of the tasks, the classifier and dynamics model are learned first using system identification as described in Section \ref{sec:sysID_learning}. Then, the cost matrices and goal conditions are learned as described in Section \ref{sec:control_param_learning}. For the eMOSAIC controller, these learned cost matrices and goal conditions are used to design independent infinite-horizon LQR controllers, one for each mode. The root mean squared error (RMSE) between the learned and ground truth values for the masses and goal configurations for each mode are summarized in Table \ref{tab:learned_params}. 
% Additional details concerning training and experiments are provided in appendix \cite{ourpaper}.
\subsubsection{Task 1: 1-object pickup}
We consider the task of picking up an object using a 2D gripper and taking it to a goal region, specified as the black circle shown in Fig. \ref{fig:sampleTrajs}a. The classifier learns to predict two modes. Mode 1 is active when the gripper is away from the object and mode 2 is active when the relative position between the gripper and the object is smaller than a learned threshold value. For this task, we observe that the classifier learns to predict contact a little after the actual impact occurs signifying imperfectness in the learned model. The masses for the gripper and object are set as 1kg and 9kg respectively in simulation. We perform multiple trials with varying initial configurations and shape parameters. As shown in Fig. \ref{fig:success_rate}, our controller with and without impact dynamics ($\Crhorho$) and eMOSAIC generalize well to new scenarios. Our controller is robust to local inaccuracies and eMOSAIC is unaffected by them as it is purely reactive. Both of these methods are successful in taking the object to the goal location in all the trials. On the other hand, performance of the predictive controller suffers because of inaccuracy in the learned classifier. The predictive controller fails to identify a contact even after the contact has actually occurred and does not switch to the next mode. We observe that the trajectories generated by eMOSAIC are non-smooth and the impact velocities are much higher than the expert's, leading to higher cost. In contrast, the resulting cost of our controller is lower than the other controllers because of its capability to predict the next mode, curving the trajectory of the gripper in the direction of the next goal early on. The cost of our controller without $\Crhorho$ is also slightly higher as it doesn't consider the discontinuous jump in velocities of the object and gripper upon contact.
 \begin{figure*}[t]
\centering
    \includegraphics[width=0.95\textwidth]{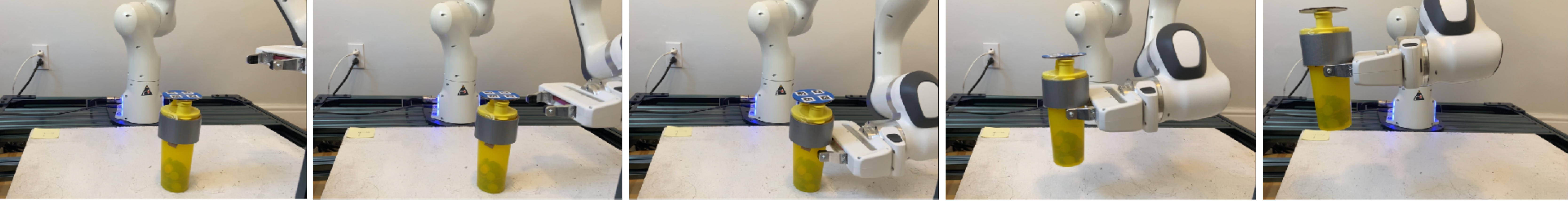}
    \vspace{-0.3cm}
    \caption{ Sequence of images showing the Franka robot arm performing a dynamic pickup task using our learned reactive and predictive controller. The controller generalizes to previously unseen locations of the object and performs the task with 80\% success rate.}
    \label{fig:robot_exp}
    \vspace{-0.5cm}
\end{figure*}
% \vspace{-0.7cm}
\subsubsection{Task 2: 2-object pickup}
We consider the task of picking up two objects using a 2D gripper and moving them to a goal region, as shown in Fig. \ref{fig:sampleTrajs}b. The model learns to identify three modes: Mode 1) relative position between the gripper and both objects is greater than a threshold value, Mode 2) relative position between the gripper and object 1 (orange) is smaller than a threshold value whereas between gripper and object 2 (blue) is larger than a threshold value, and Mode 3) relative position between the gripper and both objects is smaller than a threshold value. The masses of the gripper and objects in simulation were set as 1kg each. For this task, we observe that the learned classifier predicts contacts a little earlier than when contact actually happens. As in Task 1, we observe that our controller with and without $\Crhorho$ is robust to these local inaccuracies, eMOSAIC is unaffected by it and both generalize well to new scenarios. As shown in Fig. \ref{fig:success_rate} both these controllers were able to successfully perform the task in all trials. The success rate for the predictive controller suffers because in a large number of trials the gripper moves past the object without actually grasping it because the classifier wrongly predicted that grasp has happened. eMOSAIC performs better than our controller for this task because it is purely predictive. The inaccurate classifier leads our method to predict suboptimal trajectories in the short term, which  leads to higher cost. 
% \vspace{-0.5cm}
\subsubsection{Task 3: Pickup task with ground interactions}
This experiment evaluates the capability of our method to capture \textit{intricate behaviors} from expert demonstrations. These demonstrations were designed to depict behavior that exploits contact to reduce uncertainty. We consider the task of picking up an object placed on a table. Fig. \ref{fig:sampleTrajs}c demonstrates how the gripper first comes in contact with the table and moves along it before coming in contact with the object and pushing it along the table before picking it up. As before, we perform experiments by varying initial configurations and block sizes. We observe that the success rate for the predictive controller improves as compared to other tasks because alignment with the object before picking it up reduces uncertainty in relative positions. Our controller with and without $\Crhorho$ and eMOSAIC are successful in completing the task in every trial but the cost increases significantly if $\Crhorho$ is not considered. This is because the controller is unable to predict that the velocity of the gripper in the direction of contact goes to zero upon contact with the ground and the subsequent motion is along the ground. This leads to suboptimal predictions and thus higher cost. The cost for eMOSAIC is also high in this task. This is because the gripper and object overshoot the table due to the lack of a predictive component preparing the gripper to move up towards the goal early on, requiring a large force to correct the motion to eventually move toward the goal.
\vspace{-0.2cm}
\subsection{Robot Experiments}
We consider the task of dynamically picking up an object and taking it to a goal location as shown in Fig. \ref{fig:robot_exp}. Ten expert demonstrations were collected using human-guided kinesthetic teaching \cite{zhang2020modular} with fixed object and goal locations. In the human demonstration, the robot slows down before contact, without stopping, to prevent the object from toppling. The location of the object was tracked using April tags and a Kinect depth sensor \cite{wang2016apriltag}. The robot dynamics were linearized using Cartesian space feedback linearization. Due to a lack of reliable contact sensing, we only rely on sensing of the object location as feedback to compare the performance of our controller with the reactive controller.\\
\indent Using the method described in Section \ref{sec:sysID_learning}, the classifier learns to discretize the task into two modes. The goal condition for the first mode (coming in contact with object) was learned with RMSE of 0.015m, and the goal condition for the second mode (final configuration) was learned with a RMSE of 0.079m. To compare our controller to the reactive eMOSAIC controller we perform 10 trials with varying initial locations of object. Using our controller, the robot successfully picks up the object and takes it to the goal location 8/10 times. Average cost over successful trials with our controller is 1.1789. Using the purely reactive eMOSAIC controller, the robot successfully picks up the object and takes it to goal location 6/10 times. The average cost over the successful trials is 1.272. The sharp movements of the reactive controller right after coming in contact with the object is the main reason behind the failed attempts. Unlike in Box2D experiments, where abrupt motion in any direction can occur, such behavior does not perform well in real-world scenarios and can lead to failed grasps. As compared to eMOSAIC, our controller executes smooth trajectories before and after impact as it operates over all modes.

%% file: conclusion.tex
We presented a framework for developing a robust differentiable feedback controller for learned switching linear dynamical systems that operates over multiple dynamical modes, predicts and reacts to discontinuous mode switches and generalizes to unseen scenarios. Expert demonstrations inform mode predictions, contact dynamics, linear dynamical system parameters, cost functions and goal conditions. Experimental results show a significant improvement over eMOSAIC, a purely reactive controller, in multi-object pick-up, sliding and grasping, and real-robot grasping. We find that replanning using MPC and posterior modes improves robustness to model inaccuracies. In future work we will explore planning over modes to generalize to tasks that might require different mode sequencing. We plan to extend the capabilities of our controller to non-linear dynamical systems by learning locally linear dynamical models. 

%% file: appendix.tex
\section{APPENDIX}

\subsection{Training details}

The classifier, for all the experiments, is a single layer neural network with 8 hidden nodes. The output layer has nodes equal to the number of modes and softmax activation. The other hyperparameters used during training (system identification as well as learning control parameters) are as follows:

\begin{table}[h]
\caption{Hyperparameters}
\begin{center}
\begin{tabular}{ | c | c | } 
\hline
Optimizer & ADAM \cite{kingma2014adam} \\
\hline
Learning rate & 0.01 \\
\hline
$\beta_1$ & 0.9 \\
\hline
$\beta_2$ & 0.999 \\
\hline
$\epsilon$ & 1e-04 \\
\hline
Length of each trajectory & 12.5sec \\
\hline
$dt$ & 0.05sec \\
\hline
batch size & 10 \\
\hline
\end{tabular}
\end{center}
\label{tab:hyperparameters}
\end{table}

\subsection{Task 1: 1-object pickup}

This task involves a gripper of mass $m^g$ picking up an object of mass $m^o$ while moving in a 2D plane. The independent dynamics of the gripper and object can be written as,
\begin{gather*}
\mathbf{x}^g_{t+1} = \mathbf{A}_{t} \mathbf{x}^g_t + \mathbf{B}^g_{t} \mathbf{u}^g_t, \;\;\; \mathbf{x}^o_{t+1} = \mathbf{A}_{t} \mathbf{x}^o_t + \mathbf{B}^o_{t} \mathbf{u}^o_t \text{, where}\\
    \mathbf{x}^g_t =
    \begin{bmatrix}
    q^g_{x_t} \\
    q^g_{y_t} \\
    \dot{q}^g_{x_t} \\
    \dot{q}^g_{y_t}
    \end{bmatrix},
    \mathbf{A}_{t} = 
    \begin{bmatrix}
    1 & 0 & dt & 0 \\
    0 & 1 & 0 & dt \\
    0 & 0 & 1 & 0 \\
    0 & 0 & 0 & 1 
    \end{bmatrix}, 
    \mathbf{B}^g_{t} =
    \begin{bmatrix}
    0 & 0 \\
    0 & 0 \\
    dt/m^g & 0 \\
    0 & dt/m^g
    \end{bmatrix}\\
    \mathbf{x}^o_t =
    \begin{bmatrix}
    q^o_{x_t} \\
    q^o_{y_t} \\
    \dot{q}^o_{x_t} \\
    \dot{q}^o_{y_t}
    \end{bmatrix},
    \mathbf{B}^o_{t} =
    \begin{bmatrix}
    0 & 0 \\
    0 & 0 \\
    dt/m^o & 0 \\
    0 & dt/m^o
    \end{bmatrix},
    \mathbf{u}^g_t =
    \begin{bmatrix}
    u^g_{x_t} \\
    u^g_{y_t} 
    \end{bmatrix},
    \mathbf{u}^o_t =
    \begin{bmatrix}
    u^o_{x_t} \\
    u^o_{y_t} 
    \end{bmatrix}
\end{gather*}
$q_{x_t}, q_{y_t}$ and $\dot{q}_{x_t}, \dot{q}_{y_t}$ are positions and velocities respectively. It is important to note here that the object is unactuated ($u^o_{x_t}=0$ at all time $t$) but when grasped is affected by the force applied by the gripper. We consider two modes: Mode 1 is when gripper is free and mode 2 is when the gripper has grasped the object and is moving with it. The combined dynamics of gripper and object in different modes $\rho_t \in \{1,2\}$ is as given in \eqref{eq:dyn}, where
\begin{gather*}
    \mathbf{x}_{t} =
    \begin{bmatrix}
    \mathbf{x}^{g}_{t}\\
    \mathbf{x}^{o}_{t}
    \end{bmatrix},
    \mathbf{A}_{1_t} = \mathbf{A}_{2_t} =    \begin{bmatrix}
    \mathbf{A}_{t} & \mathbf{0}\\
    \mathbf{0} & \mathbf{A}_{t}
    \end{bmatrix}\\
    \mathbf{B}_{1_t} = 
    \begin{bmatrix}
    \mathbf{B}^g_{t}\\
    \mathbf{0}
    \end{bmatrix},
    \mathbf{B}_{2_t} = 
    \begin{bmatrix}
    \mathbf{B}^g_{t}\\
    \mathbf{B}^o_{t}
    \end{bmatrix},
    \mathbf{u}_t =
    \begin{bmatrix}
    u^g_{x_t} \\
    u^g_{y_t} 
    \end{bmatrix}
\end{gather*}
\begin{gather*}
    \mathbf{C}_{\boldsymbol{\rho}_{t-1}, \boldsymbol{\rho}_t} = \boldsymbol{\rho}_{t-1}^{\top} \boldsymbol{\rho}_t \mathbf{I} + (1-\boldsymbol{\rho}_{t-1}^{\top} \boldsymbol{\rho}_t ) \mathbf{C}\\
    C_{1,2} = C_{2,1} = \frac{1}{m^g+m^o}
    \begin{bmatrix}
    1 & 0 & 0 & 0 & 0 & 0 & 0 & 0\\
    0 & 1 & 0 & 0 & 0 & 0 & 0 & 0\\
    0 & 0 & m^g & 0 & 0 & 0 & m^o & 0\\
    0 & 0 & 0 & m^g & 0 & 0 & 0 & m^o\\
    0 & 0 & 0 & 0 & 1 & 0 & 0 & 0\\
    0 & 0 & 0 & 0 & 0 & 1 & 0 & 0\\
    0 & 0 & m^g & 0 & 0 & 0 & m^o & 0\\
    0 & 0 & 0 & m^g & 0 & 0 & 0 & m^o\\
    \end{bmatrix}
\end{gather*}
Here, we observe that there are only two parameters that we need to learn to populate the above matrix, $C_1 = \frac{m^g}{m^g+m^o}$ and $C_2 = \frac{m^o}{m^g+m^o}$. Thus, during system identification we learn the four parameters: $C_1, C_2, m^g, m^o$ in addition to the classifier.

Expert trajectories for this task are generated using the following cost parameters for the respective modes in iLQR,
\begin{gather*}
    Q_1 = Q_2 = diag([1, 1, 1, 1, 10^{-6}, 10^{-6}, 10^{-6}, 10^{-6}])\\
    R_1 =diag([1, 1]), R_2 = diag([0.5, 0.5])
\end{gather*} where \textit{diag()} represents a diagonal matrix. During training we assume that the cost matrices diagonal.

\subsubsection{Task 2: 2-object pickup}
This task considers 3 modes: Mode 1 is when the gripper is free, mode 2 is when the gripper has grasped object 1 and mode 3 is when gripper has grasped object 2. The combined dynamics \eqref{eq:dyn} can be written using:
\begin{gather*}
    \mathbf{x}_{t} =
    \begin{bmatrix}
    \mathbf{x}^{g}_{t}\\
    \mathbf{x}^{o_1}_{t}\\
    \mathbf{x}^{o_2}_{t}
    \end{bmatrix},
    \mathbf{A}_{1_t} = \mathbf{A}_{2_t} = 
    \mathbf{A}_{3_t}=
    \begin{bmatrix}
    \mathbf{A}_{t} & \mathbf{0} & \mathbf{0}\\
    \mathbf{0} & \mathbf{A}_{t} & \mathbf{0}\\
    \mathbf{0} & \mathbf{0} & \mathbf{A}_{t}
    \end{bmatrix}\\
    \mathbf{B}_{1_t} = 
    \begin{bmatrix}
    \mathbf{B}^g_{t}\\
    \mathbf{0}\\
    \mathbf{0}
    \end{bmatrix},
    \mathbf{B}_{2_t} = 
    \begin{bmatrix}
    \mathbf{B}^g_{t}\\
    \mathbf{B}^{o_1}_{t}\\
    \mathbf{0}
    \end{bmatrix},
    \mathbf{B}_{3_t} = 
    \begin{bmatrix}
    \mathbf{B}^g_{t}\\
    \mathbf{B}^{o_1}_{t}\\
    \mathbf{B}^{o_2}_{t}
    \end{bmatrix},
    \mathbf{u}_t =
    \begin{bmatrix}
    u^g_{x_t} \\
    u^g_{y_t} 
    \end{bmatrix},
\end{gather*}
In the interest of space, we write the C matrix such that it maps the velocities just before collision to the velocities just after collision, thus ignoring the rows that identically map the position before collision to position after collision.
\begin{align*}
    &C_{1,2} = C_{2,1} \\&=
    \frac{1}{m^g+m^{o_1}}
    \begin{bmatrix}
    m^g & 0 & m^{o_1} & 0 & 0 & 0\\
    0 & m^g & 0 & m^{o_1} & 0 & 0\\
    m^g & 0 & m^{o_1} & 0 & 0 & 0\\
    0 & m^g & 0 & m^{o_1} & 0 & 0\\
    0 & 0 & 0 & 0 & m^g+m^{o_1} & 0\\
    0 & 0 & 0 & 0 & 0 & m^g+m^{o_1}\\
    \end{bmatrix}\\
    &C_{2,3} =
    \frac{1}{m^g+m^{o_1}+m^{o_2}}
    \begin{bmatrix}
    m^g & 0 & m^{o_1} & 0 & m^{o_2} & 0\\
    0 & m^g & 0 & m^{o_1} & 0 & m^{o_2}\\
    m^g & 0 & m^{o_1} & 0 & m^{o_2} & 0\\
    0 & m^g & 0 & m^{o_1} & 0 & m^{o_2}\\
    m^g & 0 & m^{o_1} & 0 & m^{o_2} & 0\\
    0 & m^g & 0 & m^{o_1} & 0 & m^{o_2}\\
    \end{bmatrix}
\end{align*}
Thus, to populate each $C_{i,j}$ matrix we need 9 independent parameters. We learn these parameters along with the masses $m^g,m^{o_1},m^{o_2}$ and classifier during system identification. It is interesting to note here that we only observe $C_{1,2}$ and $C_{2,3}$ in our demonstrations for this task, hence even though we define all the other $C_{i,j}s$ as training variables they are never learned and do not affect the training time.

Expert trajectories for this task are generated using the following cost parameters for the respective modes in iLQR,
\begin{align*}
    &Q_1 = Q_2 = Q_3\\ &= diag([1, 1, 1, 1, 10^{-6}, 10^{-6}, 10^{-6}, 10^{-6}, 10^{-6}, 10^{-6}, 10^{-6}, 10^{-6}])\\
    R_1 &=diag([1, 1]), R_2 = R_3 = diag([0.5, 0.5])
\end{align*} where \textit{diag()} represents a diagonal matrix. Here again during training we assume that the cost matrices diagonal.

\subsubsection{Task 3: Pickup task with ground interactions}

This task captures four modes: Mode 1 is when the gripper moves in free space, mode 2 is when the gripper comes in contact with the table and moves along it, mode 3 is when the gripper comes in contact with the object on the table and pushes it along the table and mode 4 is when the gripper picks up the object off the table and takes it to the goal location. The dynamics for each mode can be written as,
\begin{gather*}
    \mathbf{x}_{t} =
    \begin{bmatrix}
    \mathbf{x}^{g}_{t}\\
    \mathbf{x}^{o}_{t}
    \end{bmatrix},
    \mathbf{A}_{1_t} = \mathbf{A}_{2_t} =
    \mathbf{A}_{3_t} =
    \mathbf{A}_{4_t} =
    \begin{bmatrix}
    \mathbf{A}_{t} & \mathbf{0} \\
    \mathbf{0} & \mathbf{A}_{t} 
    \end{bmatrix}\\
    \mathbf{B}_{1_t} = \mathbf{B}_{2_t}=
    \begin{bmatrix}
    \mathbf{B}^g_{t}\\
    \mathbf{0}
    \end{bmatrix},
    \mathbf{B}_{3_t} = \mathbf{B}_{4_t}=
    \begin{bmatrix}
    \mathbf{B}^g_{t}\\
    \mathbf{B}^{o}_{t}
    \end{bmatrix},
    \mathbf{u}_t =
    \begin{bmatrix}
    u^g_{x_t} \\
    u^g_{y_t} 
    \end{bmatrix},
\end{gather*}
Again, writing down the impact dynamics while ignoring the rows that identically map positions before collision to after collision,
\begin{gather*}
    C_{1,2} = 
    \begin{bmatrix}
    0 & 0 & 0 & 0\\
    0 & 0 & 0 & 0\\
    0 & 0 & 1 & 0\\
    0 & 0 & 0 & 1
    \end{bmatrix},\\
    C_{2,3} = C_{3,4} = \frac{1}{m^g+m^o}
    \begin{bmatrix}
    m^g & 0 & m^o & 0\\
    0 & m^g & 0 & m^o\\
    m^g & 0 & m^o & 0\\
    0 & m^g & 0 & m^o
    \end{bmatrix}
\end{gather*}
Thus, each of the above matrices can be populated using 4 independent parameters which we learn along with the masses $m^g, m^o$ and classifier during system identification.
The expert trajectories were generated using the following cost parameters for each mode,
\begin{gather*}
    Q_1 = diag([1, 10, 1, 1, 10^{-6}, 10^{-6}, 10^{-6}, 10^{-6}])\\
    Q_2 = Q_3 = diag([1, 1, 1, 1, 10^{-6}, 10^{-6}, 10^{-6}, 10^{-6}]),\\
    Q_4 = diag([1, 1, 0.01, 0.01, 10^{-6}, 10^{-6}, 10^{-6}, 10^{-6}]),\\
    R_1= R_2 =diag([1, 1]), R_3 = diag([0.5, 0.5]), R_4 = diag([0.1, 0.1])
\end{gather*} where \textit{diag()} represents a diagonal matrix. Here again during training we assume that the cost matrices diagonal.